\documentclass[10pt,conference]{IEEEtran}
\IEEEoverridecommandlockouts
\usepackage{amsmath,amssymb}
\usepackage{graphicx, soul}
\usepackage{graphicx}
\usepackage{booktabs} % for professional tables
\usepackage{multirow}
\usepackage{rotating}
\usepackage[colorlinks,urlcolor=blue,linkcolor=blue,citecolor=blue]{hyperref}
\usepackage[noadjust]{cite}
\usepackage{xcolor}
\usepackage{todonotes}
\usepackage[noend]{algorithmic}
\usepackage[linesnumbered,ruled,vlined,algo2e]{algorithm2e} %
\usepackage{algorithm}
\let\oldnl\nl% Store \nl in \oldnl
\newcommand{\nonl}{\renewcommand{\nl}{\let\nl\oldnl}}% Remove line number for one line
\usepackage{color}
\definecolor{red}{rgb}{1,0.2,0.2}
\definecolor{green}{rgb}{0.1,0.70,0.32}%{0.2,1,0.5}
\definecolor{blue}{rgb}{0,0,1}%{0,0,1}
\definecolor{bloo}{rgb}{0,0,1}%{0,0,1}
\definecolor{lightblue}{rgb}{0.3,0.5,1}

\newcommand{\blue}[1]{\textcolor{black}{#1}}

\definecolor{lightgray}{rgb}{0.5,0.5,0.5}
\newcommand{\ms}[2]{{#1}_{\color{lightgray}{\pm #2}}}
\usepackage[linesnumbered,ruled,vlined,algo2e]{algorithm2e} %
\newcommand{\inprod}[2]{\left< #1, #2 \right>}

\begin{document}

\title{\LARGE \bf Enhancing Robustness of Federated Learning via Server Learning}

\author{Van Sy Mai$^{1}$, Kushal Chakrabarti$^{2}$, Richard J. La$^{3}$, and Dipankar Maity$^{4}$%
 % \thanks{\textit{Corresponding author: Van Sy Mai.}} 
 \thanks{$^{1}$V. S. Mai is with National Institute of Standards and Technology (NIST), Gaithersburg, MD 20899, USA {\tt\small vansy.mai@nist.gov}}% 
 %} 
 \thanks{$^2$R. J. La is with the University of Maryland, College Park, MD 20742, USA {\tt\small hyongla@umd.edu}}%
 \thanks{$^3$K. Chakrabarti is with Tata Consultancy Services Research, Mumbai, Maharashtra 400607, India {\tt\small chakrabarti.k@tcs.com}}%
 \thanks{$^4$D. Maity is with the University of North Carolina at Charlotte, NC 28223, USA {\tt\small dmaity@charlotte.edu}}%
 %\textit{Corresponding author: Van Sy Mai.}}
 \thanks{Any mention of commercial products in this paper is for information only; it does not imply any recommendation or endorsement by NIST. U.S. Government work not protected by U.S. copyright.}}

\maketitle
\begin{abstract}
This paper explores the use of server learning for enhancing the robustness of federated learning against malicious attacks even when clients' training data are not independent and identically distributed. We propose a heuristic algorithm that uses server learning and client update filtering in combination with geometric median aggregation. We demonstrate via experiments that this approach can achieve significant improvement in model accuracy even when the fraction of malicious clients is high, even more than $50\%$ in some cases, and the dataset utilized by the server is small and could be synthetic with its distribution not necessarily close to that of the clients' aggregated data.
\end{abstract}

% \begin{IEEEkeywords}
% Federated Learning, Non-IID Data, Byzantine Attacks
% \end{IEEEkeywords}

\section{Introduction}
Federated Learning (FL) has emerged as a means of distributed learning using local data stored at clients with a coordinating server. Recent studies showed that FL can suffer from poor performance when training data at the clients are not independent and identically distributed (IID) and more severely so under Byzantine attacks, that is, malicious participants can disrupt model training by injecting faulty updates.

In general, there are two main strategies to enhance robustness of FL against Byzantine attacks. First is to use robust aggregation techniques such as coordinate-wise trimmed mean or geometric median to alleviate the impact of malicious updates \cite{blanchard2017machine,pillutla2022GeoMed}. This approach relies on a key assumption that the fraction of malicious clients  is less than 0.5 and usually a known upper bound on this fraction is available. The second approach is to use auxiliary information for filtering out potential malicious updates before the model aggregation step -- this approach relies on the quality of the auxiliary information, such as additional data available at the server \cite{xie2020zeno++}.  

Recently, \cite{mai2024study} proposed an approach for improving FL on non-IID client data using server learning (SL) even when server data is small and not necessarily close to client data distribution. Updates from malicious clients can be viewed as those from out-of-distribution and, hence, SL will have the potential to improve the overall performance.   
A natural question is the following: \textit{Can SL improve robustness of FL against Byzantine attacks}? We propose to use both SL and filtering in combination with geometric median aggregation to improve the robustness of FL on non-IID data and under Byzantine attacks. Our experiments show that such combination can achieve significant improvements in model accuracy even when the fraction of malicious clients is high and the dataset utilized by the server is small and its distribution differs from that of the clients' aggregate data.

% \subsection*{Related Work}
The proposed approach integrates two key methods to enhance the robustness of FL against Byzantine attacks, particularly in non-IID settings: the \textit{use of a server-side dataset for learning and filtering}, and the application of \textit{robust aggregation mechanisms}. This positions our work at the intersection of several active research areas in FL, as briefly outlined next.

\textbf{Byzantine-Robust Aggregation Rules:} A primary line of defense against Byzantine attacks involves developing robust aggregation rules at the server to mitigate the impact of malicious model updates.
Early and prominent methods include Krum and Multi-Krum, which select a singe client update that is closest to its neighbors \cite{blanchard2017machine}, and coordinate-wise median and trimmed mean, which are robust to outliers in individual dimensions of the model update vector. 
Our approach utilizes geometric median, a well-established robust aggregator that can tolerate up to half of the clients being malicious \cite{pillutla2022GeoMed}. More recent work has continued to refine these aggregation strategies. However, a key limitation of relying solely on aggregation is the assumption that malicious clients constitute a minority, a condition that may not hold in all scenarios. The proposed approach aims to overcome this by integrating SL.

\textbf{Filtering Malicious Updates:} Another strategy to defend against Byzantine attacks is to filter out malicious updates before the aggregation step. This can be achieved through various techniques, including anomaly detection and clustering-based approaches that identify and exclude suspicious updates \cite{xie2019zeno}. Another approach in \cite{zhang2022fldetector} aims to detect malicious clients based on their model-updates consistency, which could be ineffective against data poisoning, especially in non-IID settings. The study in \cite{xie2020zeno++} assumes that server uses a small dataset with the same distribution as the global training data for filtering. Some methods leverage similarity metrics, such as \textit{cosine similarity} between client updates and a reference model to identify and discard malicious contributions \cite{awan2021contra}. Our work explores two filtering techniques: an \textit{angle-based filter}, which is a form of similarity check against server model's gradient, and a \textit{loss-based filter} inspired by prior work, which assesses a client's update based on the server's loss function.

\textbf{Federated Learning with Server-Side Data:} The concept of leveraging a dataset at the server is a promising direction for improving FL, especially in non-IID environments. A server-side dataset can be used for various purposes, including regularization, knowledge distillation, and as a reference for detecting malicious behavior \cite{li2024federated}. In some approaches, the server is treated as a special client that participates in the learning process. In contrast to work that assumes the server data has the same distribution as the client data, we consider a more realistic scenario where the server's dataset can be small, synthetic, or have a different distribution. It builds on the idea that even a small, imperfect dataset at the server can guide the global model toward a better solution and aid in identifying harmful updates. This is particularly relevant in the context of Byzantine attacks, where malicious updates can be viewed as extreme cases of out-of-distribution data.

The novelty of the presented work lies in the synergistic combination of these three areas. By using the server's model not just for regularization but as an active component in filtering malicious updates, the proposed method can achieve an ``honest majority" condition {even when more than half of the participating clients are malicious}. This integration of server learning with robust aggregation and filtering presents a \textit{complementary} approach to enhancing the security and reliability of FL in the face of Byzantine threats.

The rest of the paper is organized as follows. The problem formulation and our proposed approach are given in Section~\ref{sec_prob}. The main algorithm is described in Section~\ref{sec_main_algorithm}, followed by experimental evaluations in Section~\ref{sec_experiment}. We conclude in Section~\ref{sec_conclusion}.

\textbf{Notation}: For an integer $n>0$, let $[n]:=\{1, \ldots, n\}$. For a finite set $\mathcal{D}$, $|\mathcal{D}|$ denotes its cardinality. For $x{\in} \mathbb{R}^n$, $\|x\|$ denotes its 2-norm. The inner product of $x$ and $y$ is denoted by $\langle x, y \rangle$.

\section{Problem Formulation and Our Approach}\label{sec_prob}

\subsection{Problem Formulation}
    \label{subsec:Formulation}

The goal of FL is to minimize the total loss of a model over training data distributed at multiple clients under the coordination of a server. 
Consider $N$ clients with their datasets denoted by $\{\mathcal{D}_i\}_{i=1}^N$. For each client $i \in [N]$, let  $f_i(x) := \frac{1}{n_i} \sum_{s\in \mathcal{D}_i} \ell(x, s)$ with $n_i{=} |\mathcal{D}_i|$ be the average sample loss on the local dataset $\mathcal{D}_i$ under model $x$ and sample loss function $\ell$. We are interested in finding a model that minimizes the weighted average loss over all clients: 
\begin{align} 
x^* \in \textstyle \arg\min_{x \in \mathbb{R}^d} \ \big( F(x) =\sum_{i\in [N]} p_i f_i(x) \big), 
    \label{eqProblem_normal}
\end{align}
where $(p_i)_{i \in [N]}$ be a probability vector, which is usually given by $p_i = \frac{n_i}{n}$ for all $i\in [N]$ with $n = \sum_{i \in [N]} n_i$.%\todo[inline]{\small is $n = \sum_{i \in [N]} n_i$?} 

The main role of the server is to coordinate clients' learning, but it can also incorporate additional regularization  as follows: 
\begin{align}
 \min_{x\in \mathbb{R}^d} ~~ \gamma f_0(x) + F(x) = \gamma f_0(x) + \textstyle \sum_{i\in [N]} p_i f_i(x), 
    \label{eqProblem_SL_normal}
\end{align}
where $f_0$ is the regularizer. We are interested in the case where a fraction of clients, denoted by $\mathcal{B}$, are malicious and their goal is to prevent honest clients from learning a good model by injecting bad updates in the FL process. Additionally, the server has access to some related data, denoted by $\mathcal{D}_0$, which could be synthetic or real data collected either from test clients or beta users, and uses it for learning with a loss function $f_0(x) := \frac{1}{n_0} \sum_{s \in \mathcal{D}_0} \ell(x, s).$ 
Thus, our goal becomes
% {\small 
\begin{align}
 \min_{x\in \mathbb{R}^d} ~~ \gamma f_0(x) + F'(x):= \gamma f_0(x) + \textstyle \sum_{i\in [N]\setminus \mathcal{B}} p_i f_i(x).
    \label{eqProblem_SL_Byzantine}
\end{align}
% }

\subsection{Proposed Approach}
Our idea is to use server data not just for filtering but also for learning to enhance the robustness of FL. This can be achieved by noting that when the distribution of server data $\mathcal{D}_0$ and that of honest clients $\mathcal{D}'=\cup_{i\in [N]\setminus\mathcal{B}}\mathcal{D}_i$ are close, server's loss function $f_0$ will be similar to the overall loss function $F'$ in \eqref{eqProblem_SL_Byzantine}. Thus, when the current model is far from an optimal point, the gradient $\nabla f_0$ will more or less track the global gradient $\nabla F$, even when individual clients' gradients $\nabla f_i$ including malicious ones do not follow $\nabla F$ closely. This fact can be employed in two ways.
First, updates from clients that significantly increase or do not improve server's loss $f_0$ can be regarded as unuseful or malicious and thus should be filtered out. Second, when the updated model obtained by aggregating clients' updated models does not make (much) progress, $\nabla f_0$ will help improve the updated model. In fact, as shown in~\cite{mai2024study}, significant improvements can still be achieved even when the distributions of $\mathcal{D}_0$ and $\mathcal{D}'$ are not very similar as long as their difference is small in relation to the divergence in the distributions of clients' datasets; %the non-IID natureness of clients' data; 
this will likely be the case in the presence of malicious clients. More importantly, by capitalizing on server as an honest learner and employing incremental server filtering and learning, we can practically achieve ``honest majority'' even when more than half of clients are malicious. Our approach can be viewed as a combination of \cite{mai2024study} and \cite{xie2020zeno++}.

\section{Main Algorithm} \label{sec_main_algorithm}

Lines 1--7 of Algorithm~\ref{RoFSL} are similar to the FedAvg algorithm \cite{McMahan2017}, where in each global round $t$, each selected client~$i$, if not a malicious one, (i) receives the current global model $x_t$ from the server, (ii) performs $K$ steps of the Stochastic Gradient Descent (SGD) algorithm using its local data $\mathcal{D}_i$ (\textsc{LocalSGD}) with learning rate $\eta_l$, and (iii) returns to the server its latest update. On the other hand, if client $i$ is malicious, then the reported update can be anything.

The server then combines all the updates using a robust aggregation scheme and uses the resulting updated model to learn locally by performing $K_0$ steps of \textsc{LocalSGD} with learning rate $\eta_0$. 
%Clearly, FSL has the same computation and communication costs at the clients as the FedAvg algorithm. 
We now describe these steps in detail below.

\begin{algorithm2e}[tb]
\small
\caption{\small \textsc{RoFSL}: Robust FL via Server Learning} \label{RoFSL}
\DontPrintSemicolon
% \begin{algorithmic}%[1]
\nonl\textbf{Server:}{initial $x_0$, learning rates $\eta_g, \eta_0$, no.\! steps $K_0$, weight $\gamma$}\!\;
\nonl\textbf{Clients:} \blue{learning rate $\eta_l$, no. steps $K$}\;
% \nonl\textbf{Server:} %\qquad\text{ Combine FL and Local Learning}
% init. $x_0$, $K$, $K_0$, $\eta_l, \eta_g, \eta_0, \gamma$\;
\For{$t = 0, \ldots, T-1$}{
    sample a subset $\mathcal{S}$ of clients\;% (with $|\mathcal{S}| = S$)\;
    broadcast $x_t$ to clients in $\mathcal{S}$\;
    \ForAll{$\mathrm{clients}~i \in \mathcal{S}$}{
        $x^{(i)}_{t} \gets \begin{cases}
            \texttt{LocalSGD}(f_i, x_t, \eta_l, K), \quad &i\notin \mathcal{B}\\
            \texttt{Attack} &i \in \mathcal{B}
        \end{cases}$\;
        upload $x^{(i)}_{t}\to$ Server\; 
    }
    % $\Delta_t \gets \frac{\sum_{i\in \mathcal{S}}\Delta^{(i)}_t}{|\mathcal{S}|}$\;
    % $\mathcal{S}_0 \gets \textsc{Filter}(\mathcal{S})$\; 
    % $\bar{x}_t \gets x_t + \eta_g \Big(\frac{\sum_{i\in \mathcal{S}_0}x^{(i)}_{t,K}}{|\mathcal{S}_0|} -x_t \Big)$\;
    $\bar{x}_t \gets \texttt{RobustAggr}\big(x^{(i)}_{t}, i\in \mathcal{S} \big)$\;
    %$\bar{x} = \frac{1}{n} \sum_{i=1}^n x^{i}_{t+1}$\;
    % $z_{t+1} = \textsc{LocalSGD}( \bar{x}_t, \gamma\eta_0, K_0,  \mathcal{D}_0)$\;
    % $x_{t+1} = z_{t+1} + \beta(x_t - x_{t-1})$\qquad\blue{Momentum, added 221201}\; 
    % %$x_{t+1} = (1-\gamma)\times \textsc{LocalSGD}( \bar{x}_t, K_0, \eta_0, \mathcal{D}_0) + \gamma\times x_t$\;
    $x_{t+1} \gets \texttt{LocalSGD}(\gamma f_0, \bar{x}_t, \eta_0, K_0)$\;
}
\vspace{-6pt}
\nonl\;
\nonl$\texttt{LocalSGD}(f, x, \eta, K)$:\; %\qquad \text{ \righttriangle Local learning via SGD}
    $y_0 \gets x$\;
    \For{$k=0,\ldots,K-1$}{
        %compute a minibatch gradient $g(y_k)$\;
        $g(y_k)\gets$ unbiased estimate of $\nabla f(y_k)$\;
        $y_{k+1} \gets y_{k} - \eta g(y_k)$\;
    }
    \textbf{return}: $y_K$
% \end{algorithmic}
\end{algorithm2e}

\subsubsection{Robust Aggregation}
As we mentioned above, when far from convergence, the gradient $\nabla f_0$ will more or less track the global gradient $\nabla F$. This fact can be employed to filter out updates from malicious clients using the following ideas.

% \begin{itemize}
%     \item 
    \paragraph{Angle-based filtering ({\normalfont\texttt{AF}})}: Client updates $\Delta x_t^{(i)}$ that are too far from the direction of $-\nabla f_0(x_t)$ deem unuseful or likely malicious. This can be done by using a threshold $\alpha \in [0,1]$ on their cosine similarity, defined as $\mathrm{cos\_sim}(x, y) = \frac{\inprod{x}{y}}{\|x\|\|y\|} 
    $ for any $x, y \in \mathbb{R}^d$. For $\mathcal{S} \subset [N]$, define
    $$
    \texttt{AF}_{\alpha}(\mathcal{S}) := \{i\in \mathcal{S}:\quad \mathrm{cos\_sim}(\Delta x^{(i)}, -\nabla f_0(x)) \ge \alpha \}. 
    $$
    Setting a suitable threshold requires a careful consideration because a large $\alpha$ will result in less (even zero) contributions from benign clients whereas a small $\alpha$ will likely allow malicious updates. Additionally, as this filter only looks at the direction but not the magnitude of the update, it could happen that malicious clients can provide large updates near the cosine similarity threshold to significantly slow down or disrupt the convergence. For these reasons, we believe that this filter should use a rather loose threshold such as $\alpha=0$ and be used together with additional robust aggregation steps to be effective. 
    
    % \item 
    \paragraph{Loss-based filtering (\texttt{LF})} We adapt the stochastic descent score in \cite{xie2020zeno++} as follows: 
    $$
    \mathrm{sc}_{\rho}^{(i)}(x) = -\langle\Delta x^{(i)}, \nabla f_0(x)\rangle - \rho \|\Delta x^{(i)}\|^2, 
    $$
    which can be viewed as a second-order approximation to $f_0(x) - f_0(x+\Delta x^{(i)})$, i.e., the improvement of the server loss when using client $i$'s model. Clearly, this score is related to the cosine similarity in \texttt{AF} through the inner product term. But, it also takes into account the magnitude of the update penalizing large deviations via $- \rho\|\Delta x^{(i)}\|^2$. In \cite{xie2020zeno++}, the authors consider $\Delta x^{(i)}$ to be (negative of) the stochastic gradient of client $i$ (normalized to have the same norm as $\nabla f_0(x)$) and use a fixed threshold $\epsilon$ to filter clients' models based on their scores, e.g., $\mathrm{sc}_{\rho}^{(i)}(x) \ge -\epsilon$. Again, choosing a suitable threshold $\epsilon$ (even just the range of it) is nontrivial as the fraction of malicious clients and the non-IIDness of client data are unknown. In our case, we allow $\Delta x^{(i)}$ to be the client update (after a number of local training steps without normalizing) and consider their scores as defined above for filtering. In general, one could consider clustering the scores, but for simplicity, we filter out a fixed fraction of updates with the lowest scores, denoted by $\theta \in (0,1)$, in every round. Specifically, 
    for any set $\mathcal{S}$, we first sort   $\mathrm{sc}_{\rho}^{(i_1)}(x) \le \mathrm{sc}_{\rho}^{(i_2)}(x)\le {\ldots}\le \mathrm{sc}_{\rho}^{(i_{|\mathcal{S}|})}(x)$ and then find 
    $$
    \texttt{LF}_{\rho, \theta}(\mathcal{S}) = \{i_{\lceil \theta\times|\mathcal{S}|\rceil}, \ldots, i_{|\mathcal{S}|}\} \subset \mathcal{S}
    $$
    as the set of accepted clients. 
    For example, choosing $\theta=0.5$ removes half of the sampled clients in each round if we believe that the majority of clients is honest. Of course, it is unlikely that the fraction of malicious clients is known in advance and this assumption may not hold. Furthermore, even if the fraction of malicious clients is known and less than 0.5, since $\mathcal{S}$ is randomly selected in each round, the majority of sampled clients in a round can be malicious. Thus, a robust aggregation step is still needed after this filtering step. 

    % \item 
    \paragraph{Model aggregation} Several robust aggregation schemes have been employed recently in distributed and federated learning. In this paper, we consider geometric median as it has been shown to be relatively robust at the cost of higher computational complexity. In fact, although 
    $$
    \texttt{GeoMed}(x_i, i\in \mathcal{S}) \in \textstyle \arg\min_x   \ \sum_{i\in \mathcal{S}} \|x-x_i\|
    $$
    has no closed form solution, an approximate solution can be computed efficiently using numerical methods such as a Weiszfeld-type algorithm \cite{pillutla2022GeoMed} or an interior-point method~\cite{cohen16GeoMed_lineartime}. 

    % \item 
    \paragraph{Norm clipping} Even with filtering and robust aggregation steps in place, the aggregated model from clients in a round can still be affected by malicious updates if they are the majority in the current round. To limit their potential impact, we use a clipping step after model aggregation: 
    $$
    \texttt{Clip}_{\tau}(x) = \min\big(1, {\tau}/{\|x\|}\big) \times x, 
    $$
    where $\tau>0$ is a norm clipping parameter set by the server. This is different from gradient clipping in local training steps of the clients. It also presents a similar trade-off for robustness as setting a too conservative $\tau$ would also limit the contribution of honest clients. Our aim is to use server learning to offset such limit. 
% \end{itemize} 
As a result, line 7 of our algorithm (RoFSL in Algorithm 1) performs the following: 
\[
\texttt{RobustAggr} \equiv \texttt{Clip}_{\tau}\circ \texttt{GeoMed}\,\circ\,\texttt{Filter}, 
\]
where $\texttt{Filter}$ can be (i) $\texttt{AF}_{\alpha}$, (ii)  $\texttt{LF}_{\rho, \theta}$, or \texttt{0F} (neither).

\subsubsection{Server Learning} In the case without attacks, it has been shown in \cite{mai2024study} that SL can help improve convergence significantly. In the presence of Byzantine clients, we believe the role of SL is even more significant as it can be viewed as an honest learner providing (i) direction for update filtering as shown above and (ii) enhancement to model convergence. More importantly, by employing incremental SL, we can practically achieve ``honest majority'' condition without requiring more than half of the clinents to be honest. Since the server data can be small and different from the aggregated data distribution of honest clients, we do not want to over-utilize SL to avoid overfitting server data; this can be achieved by tuning the weight $\gamma$ in \eqref{eqProblem_SL_Byzantine}. Additionally, we also impose the same norm-clipping step at the server to further limit such overfitting while providing enough room for the server's update to act as correction to the aggregated model from clients. Finally, SL in \cite{mai2024study} adopts a ``pseudo-gradient" step with a step size of $\eta_g>1$ using the averaged  update from clients prior to the server learning steps, namely, 
$$\bar{x}_t \gets x_t + \eta_g \textstyle \sum_{i\in \mathcal{S}} (x^{(i)}_{t} -x_t)/|\mathcal{S}|.$$
In Algorithm~\ref{RoFSL}, we ignore this ``pseudo-gradient" step (by setting $\eta_g = 1$ and using \texttt{RobustAggr} described above) because we believe it can potentially amplify the effect of malicious updates and reduce the robustness of the algorithm. We will illustrate this point in our experiments below. %\vspace{10 pt}

\section{Experimental Results}
    \label{sec_experiment}

We now illustrate the benefits of RoFSL  through experiments using EMNIST \cite{cohen2017emnist} and CIFAR-10 \cite{Krizhevsky2009}. We evaluate robustness under non-IID settings while varying the fraction of Byzantine clients that  perform an equal mix of sign-flipping and label-flipping attacks. Sign-flipping directly corrupts model updates, stressing the geometric robustness of the aggregation rule to adversarial outliers. Label-flipping, in contrast, represents a data-poisoning attack whose biased updates are harder to distinguish from honest ones in the presence of non-IID data. By combining these complementary attack types and varying the Byzantine fraction, we obtain a comprehensive and realistic assessment of robustness across both data heterogeneity and adversarial intensity.

\subsection{Setup}

\subsubsection{Data} 
For CIFAR-10, we use 50k samples for training and 10k samples for testing. For EMNIST, we use a balanced dataset with 45 label classes, 108k samples for training and 18k for testing. The training data is split among $N$ clients using Dirichlet distributions. 

\begin{figure}[!tb]
\centering
\includegraphics[width=0.45\linewidth]{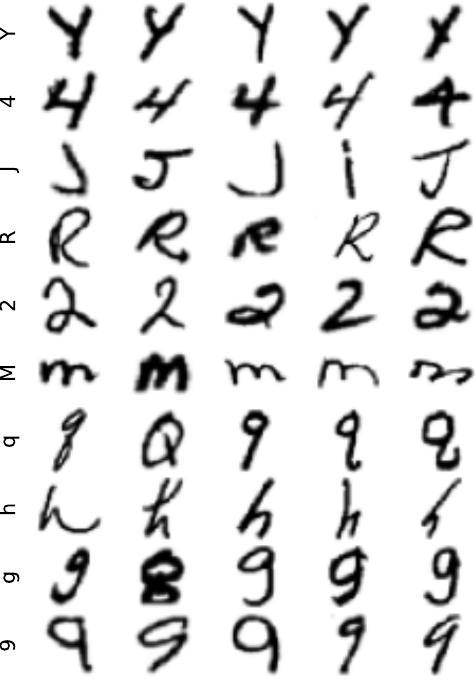}
\hfill 
\vline
\hfill 
\includegraphics[width=0.425\columnwidth]{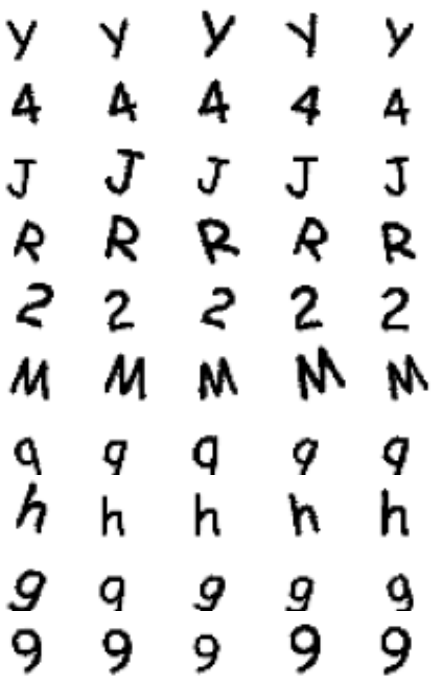} 
    \caption{Left: EMNIST training examples. Right: Server's synthetic examples.}
    \label{fig_synthetic_emnist}
\end{figure}

\begin{figure}[!t]
\centering
\includegraphics[width=0.46\linewidth]{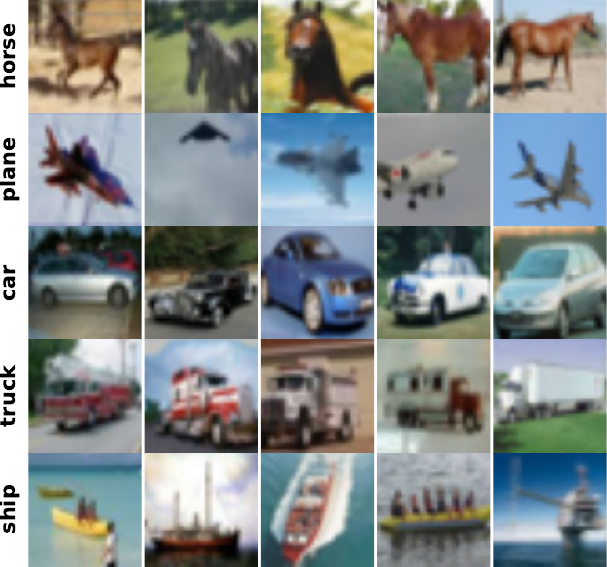}
\hfill 
\includegraphics[width=0.44\columnwidth]{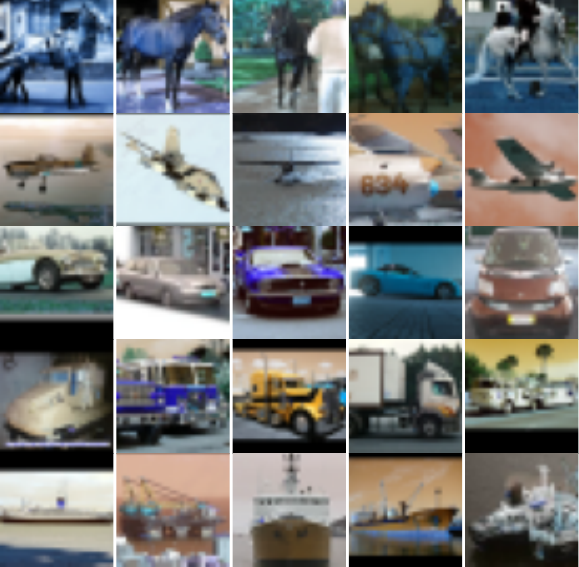}
    \caption{Left: CIFAR-10 training examples. Right: Server's STL-10 examples.}
    \label{fig_synthetic_cifar}
    \vspace{-3mm}
\end{figure}

For server, we consider synthetic data or data from a different source. Specifically, for EMNIST, we provide the server $n_0 = 900$ \textit{synthetic} examples by generating for each label class $20$ images of the corresponding letter or number using a cursive font with 5 rotation angles $\{-20, -10, 0, 10, 20\}$ and 4 sizes;\footnote{We first plot each character or number in a 2 inch $\times$ 2 inch figure using  font sizes $\{100, 110, 120, 130\}$ in points with each point equal to $1/72$ inch, and then resize it to a 28 pixel $\times$ 28 pixel figure.} see Fig.~\ref{fig_synthetic_emnist} for a comparison of this synthetic data and EMNIST. Here, we note that many classes in the balanced EMNIST dataset include both upper-case and lower-case letters whereas the synthetic data at the server only contains a single case for each class (e.g., \texttt{J} and \texttt{M} shown in Fig~\ref{fig_synthetic_emnist}). 
For CIFAR-10, we collect $n_0 = 900$ images from the dataset STL-10 with $9$ similar label classes as in CIFAR-10, each with $100$ examples;\footnote{STL-10 images were acquired from labeled examples on ImageNet; data available at: https://cs.stanford.edu/$\sim$acoates/stl10/} see Fig.~\ref{fig_synthetic_cifar} for an illustration of this data, and note that the class \texttt{frog} is absent in STL-10.


In both cases, the server data exhibits  shortcomings of synthetic data that one often expects in practice, namely limited quantity, different distributions, and possibly missing data or partially correct labeled data. We do not attempt to improve the quality and quantity of server data; our goal is to examine the benefits of SL when it is performed on data with a significantly different distribution than that of clients' data.

\subsubsection{Models}
% Similar to \cite{McMahan2017, Karimireddy2020}, 
We use simple neural networks with 2 convolutional layers (with a kernel size of 3 and  {\small\verb|padding=1|}) and 2-3 dense layers. All layers use \texttt{ReLU} as activation functions except for the last dense layer that uses \texttt{softmax}. The detailed structures are as follows. For \textit{EMNIST} dataset, we use {\small Conv2d(1,32)--Conv2d(32,64)--MaxPool2d(2)--Dropout(0.25)--Dense(128)--Dropout(0.5)--Dense(45)}, and for \textit{CIFAR-10}, {\small Conv2d(3,32)--MaxPool2d(2)--Conv2d(32,64)--MaxPool2d(2)--Dropout(0.25)--Dense(128)--Dropout(0.5)--Dense(10).} 
These models are by no means the state-of-the-art, but are enough for our purpose of illustrations and comparisons. 
% We use cross-entropy as the loss function in training. 

\subsubsection{Attacks}
In each experiment, we assign a fraction $\beta$ of the clients (selected uniformly at random without replacement) to be attackers and consider an equal mix of two attack types, namely sign-flipping and label-flipping, denoted by $\mathcal{B}_s$ and $\mathcal{B}_l$, respectively. This ensures that the robust aggregation method is not tuned or accidentally specialized to a single type.
%We use the same number of attacks for each type, i.e., $|\mathcal{B}_s| = |\mathcal{B}_l|$.  

\paragraph{Sign-flipping}  Attackers train their model as usual to obtain $\Delta x_t^{(i)}$ but send server their update with signs flipped $\Delta \check{x}_t^{(i)} = - \nu_i\Delta x_t^{(i)}$, or equivalently, $\check{x}_t^{(i)} = x_t - \nu_i\Delta x_t^{(i)}$, where $\nu_i >0$ controls the strength of the attack selected uniformly at random in $[0.1, 10.1]$ for all $i\in \mathcal{B}_s$. 

\paragraph{Label-flipping} Attackers shift their (encoded) labels by $1$ (i.e., class $c \to c+1$) and scale their learning rate by a factor $\nu_i$ selected uniformly at random in $[0.1, 2.1]$ for every $i\in \mathcal{B}_l$. In each round, attackers train their model and send the server the corresponding update. 

% \paragraph{Coordinated-attack} The attackers coordinate among themselves to design the attack vectors. Here we consider a coordinated sign-flipping attacked. 
% At each round, the selected attackers $i \in \mathcal{S}_t \cap \mathcal{B}_c$ compute their own increment $\Delta x_t^{(i)}$ as usual and then find the sign-flipped vectors according to the sign-flipping attack: $\Delta \check{x}_t^{(i)} = - \nu_i\Delta x_t^{(i)}$. 
% Afterwards, they `agree' on a coordinated attack vector $\Delta \check{x}^{\text{group}}_t = {\tt{aggregate}}(\Delta \check{x}_t^{(i)}; i\in \mathcal{S}_t \cap \mathcal{B}_c)$.\todo{\small what would be a good aggregation function? GeoMed?} 
% Each attacker then send the a noisy version of this vector $\Delta \check{x}^{\text{group}}_t + \varepsilon^{(i)}_t$ to the server. 
% The noise level is small to ensure their attack vectors are aligned, but not too small so that the server can easily detect the collusion. 

%%%%%%%%%%%%%%%%%%%%%%%%%%%%%%%%%%%%%%%%%%%%%%%
% Python file: C:\Users\vnm5\OneDrive - NIST\Documents\Enki\RoFSL-clip\report_emnist_byz.py
% fdir = "save/250221R1/EMNIST_Dir0.1_N450_E2_B50_lr0.1_S20/"
% fdir = "save/250219R1/EMNIST_Dir0.3_N450_E2_B50_lr0.1_S20/"
% #%% BIG TABLE
%%%%%%%%%%%%%%%%%%%%%%%%%%%%%%%%%%%%%%%%%%%%%%%
\begin{table*}[!t]
\caption{Accuracies after 500 rounds for EMNIST and 1500 rounds for CIFAR10 with different parameters of Server Learning $\gamma$, Byzantine fraction $\beta$, and Filtering: no filter (\texttt{0F}),  Angle-based filter (\texttt{AF}), and Loss-based filter (\texttt{LF}).}
\label{table_acc_emnist}
\vspace{-0.2in}
\begin{center}
% \begin{small}
% \begin{sc}
\resizebox{\linewidth}{!}{%
% \vskip -0.1in
\begin{tabular}{c|c||c|c|c||c|c|c||c|c|c}
\toprule
\multicolumn{2}{c||}{EMNIST}%{\texttt{Dir$0.1$}}
&\multicolumn{3}{c||}{$\mathbf{\beta=0}$}
&\multicolumn{3}{c||}{$\mathbf{\beta=0.3}$}
&\multicolumn{3}{c}{$\mathbf{\beta=0.6}$}\\
% \midrule
\multicolumn{2}{c||}{\texttt{SL $\gamma$}}
& \texttt{0F} & \texttt{AF} & \texttt{LF} 
& \texttt{0F} & \texttt{AF} & \texttt{LF} 
& \texttt{0F} & \texttt{AF} & \texttt{LF}\\
\midrule
\multirow{7}{*}{\begin{turn}{90} EMNIST ~\texttt{Dir$(0.1)$~} \hfill \end{turn}}
&$0.0$
& $\ms{84.96}{0.18}$ & $\ms{83.46}{0.57}$ & $\ms{84.11}{0.26}$ 
& $\ms{81.47}{0.58}$ & $\ms{69.12}{4.33}$ & $\ms{67.07}{4.64}$ 
& $\ms{4.78}{2.02}$ & $\ms{14.31}{3.53}$ & $\ms{2.22}{0.00}$

\\[3pt]

&$.05$
& $\ms{84.89}{0.15}$ & $\ms{84.28}{0.17}$ & $\ms{83.70}{0.35}$ 
& $\ms{81.68}{0.50}$ & $\ms{82.25}{0.24}$ & $\ms{83.45}{0.18}$ 
& $\ms{7.53}{2.69}$ & $\ms{29.03}{3.90}$ & $\ms{74.41}{2.15}$ 
\\[3pt]

&$0.1$
& $\ms{84.91}{0.11}$ & $\ms{84.10}{0.21}$ & $\ms{83.45}{0.29}$ 
& $\ms{81.74}{0.42}$ & $\ms{82.57}{0.26}$ & $\ms{83.34}{0.14}$ 
& $\ms{7.80}{2.77}$ & $\ms{37.67}{4.50}$ & $\ms{77.32}{1.51}$ 
\\[3pt]

&$0.2$
& $\ms{84.82}{0.17}$ & $\ms{84.00}{0.18}$ & $\ms{83.38}{0.28}$ 
& $\ms{81.61}{0.44}$ & $\ms{82.08}{0.24}$ & $\ms{83.29}{0.27}$ 
& $\ms{8.01}{2.82}$ & $\ms{48.94}{4.20}$ & $\ms{76.93}{2.02}$ 
\\[3pt]

&$0.5$
& $\ms{84.83}{0.12}$ & $\ms{83.89}{0.26}$ & $\ms{83.17}{0.19}$ 
& $\ms{81.47}{0.33}$ & $\ms{81.34}{0.37}$ & $\ms{83.08}{0.25}$ 
& $\ms{8.91}{2.82}$ & $\ms{42.89}{2.73}$ & $\ms{76.44}{2.36}$ 
\\[3pt]

&$1.0$
& $\ms{84.78}{0.11}$ & $\ms{83.68}{0.25}$ & $\ms{83.17}{0.25}$ 
& $\ms{81.53}{0.35}$ & $\ms{80.60}{0.41}$ & $\ms{83.02}{0.24}$ 
& $\ms{9.11}{3.02}$ & $\ms{41.31}{2.10}$ & $\ms{75.91}{2.66}$ 
\\[3pt]

&$2.0$
& $\ms{84.78}{0.11}$ & $\ms{83.62}{0.25}$ & $\ms{83.18}{0.28}$ 
& $\ms{81.44}{0.37}$ & $\ms{80.79}{0.36}$ & $\ms{83.05}{0.20}$ 
& $\ms{8.94}{3.03}$ & $\ms{39.54}{2.65}$ & $\ms{75.96}{2.47}$ 
\\[3pt]

% \midrule
\midrule
\multirow{7}{*}{\begin{turn}{90} EMNIST ~\texttt{Dir$(0.3)$~}  \hfill \end{turn}}
& $0.0$
& $\ms{85.75}{0.14}$ & $\ms{84.07}{0.44}$ & $\ms{84.93}{0.26}$ 
& $\ms{82.95}{0.48}$ & $\ms{71.07}{8.24}$ & $\ms{71.40}{6.77}$ 
& $\ms{4.14}{3.76}$ & $\ms{19.03}{8.84}$  & $\ms{2.22}{0.00}$ 
\\[3pt]

& $.05$
& $\ms{85.74}{0.12}$ & $\ms{85.13}{0.25}$ & $\ms{84.92}{0.15}$ 
& $\ms{82.85}{0.37}$ & $\ms{83.73}{0.28}$ & $\ms{84.67}{0.19}$ 
& $\ms{5.88}{4.12}$ & $\ms{30.91}{7.58}$ & $\ms{20.73}{6.22}$ 
\\[3pt]

& $0.1$
& $\ms{85.65}{0.10}$ & $\ms{85.03}{0.26}$ & $\ms{84.74}{0.20}$ 
& $\ms{82.87}{0.31}$ & $\ms{83.67}{0.25}$ & $\ms{84.56}{0.18}$ 
& $\ms{6.01}{3.96}$ & $\ms{31.69}{9.02}$ & $\ms{74.66}{4.55}$ 
\\[3pt]

& $0.2$
& $\ms{85.63}{0.10}$ & $\ms{84.29}{0.32}$ & $\ms{84.61}{0.15}$ 
& $\ms{82.94}{0.25}$ & $\ms{81.70}{0.41}$ & $\ms{84.53}{0.16}$ 
& $\ms{6.37}{3.89}$ & $\ms{52.90}{5.92}$ & $\ms{79.86}{1.85}$ 
\\[3pt]

& $0.5$
& $\ms{85.52}{0.10}$ & $\ms{68.39}{0.42}$ & $\ms{84.55}{0.13}$ 
& $\ms{82.82}{0.22}$ & $\ms{35.05}{0.24}$ & $\ms{84.37}{0.14}$ 
& $\ms{6.52}{3.89}$ & $\ms{20.53}{0.00}$ & $\ms{79.22}{2.27}$ 
\\[3pt]

& $1.0$
& $\ms{85.49}{0.09}$ & $\ms{61.11}{0.15}$ & $\ms{84.48}{0.09}$ 
& $\ms{82.80}{0.24}$ & $\ms{35.56}{0.00}$ & $\ms{84.39}{0.14}$ 
& $\ms{6.75}{4.07}$ & $\ms{20.21}{0.00}$ & $\ms{79.59}{2.47}$ 
\\[3pt]

& $2.0$
& $\ms{85.48}{0.08}$ & $\ms{82.92}{0.23}$ & $\ms{84.52}{0.16}$ 
& $\ms{82.73}{0.23}$ & $\ms{52.70}{0.34}$ & $\ms{84.33}{0.14}$ 
& $\ms{6.60}{4.00}$ & $\ms{23.87}{0.00}$ & $\ms{78.79}{2.62}$ 
\\[3pt]

% \bottomrule
% \end{tabular}%
% }
% \vskip 0.1in
% % % \end{sc}
% % % \end{small}
% % \end{center}
% % % \vskip -0.25in
% % \end{table*}
% % %%%%%%%%%%%%%%%%%%%%%%%%%%%%%%%%%%%%%%%%%%%%%%%

% % %%%%%%%%%%%%%%%%%%%%%%%%%%%%%%%%%%%%%%%%%%%%%%%
% % % Python file: C:\Users\vnm5\OneDrive - NIST\Documents\Enki\RoFSL-clip\report_emnist_byz.py
% % % #%% BIG TABLE
% % %%%%%%%%%%%%%%%%%%%%%%%%%%%%%%%%%%%%%%%%%%%%%%%
% % \begin{table*}[!t]
% % \caption{Accuracy after 1500 rounds for CIFAR10 with different parameters of Server Learning $\gamma$, Byzantine fraction $\beta$ and Filtering; \texttt{0F}: no filter,  \texttt{AF}: Angle-based filter,  \texttt{LF}: Loss-based filter.}
% % \label{table_acc_cifar}
% % \vspace{-0.2in}
% % \begin{center}
% % \begin{small}
% % \begin{sc}
% \resizebox{\linewidth}{!}{%
% % \vskip -0.1in
% \begin{tabular}{c|c||c|c|c||c|c|c||c|c|c}
% \toprule
% \multicolumn{2}{c||}{CIFAR-10}%{\texttt{Dir$0.1$}}
% &\multicolumn{3}{c||}{$\mathbf{\beta=0}$}
% &\multicolumn{3}{c||}{$\mathbf{\beta=0.3}$}
% &\multicolumn{3}{c}{$\mathbf{\beta=0.6}$}\\
% % \midrule
% \multicolumn{2}{c||}{\texttt{SL $\gamma$}}
% & \texttt{0F} & \texttt{AF} & \texttt{LF} 
% & \texttt{0F} & \texttt{AF} & \texttt{LF} 
% & \texttt{0F} & \texttt{AF} & \texttt{LF}\\
\midrule 
\midrule
\multirow{7}{*}{\begin{turn}{90} CIFAR-10~\texttt{Dir$(0.1)$~}  \hfill \end{turn}}
&$0.0$
& $\ms{71.40}{0.60}$ & $\ms{61.87}{2.24}$ & $\ms{67.38}{1.13}$ 
& $\ms{61.36}{1.51}$ & $\ms{43.63}{3.61}$ & $\ms{52.03}{2.85}$ 
& $\ms{11.78}{1.50}$ & $\ms{11.07}{1.51}$ & $\ms{11.81}{0.95}$ 
\\[3pt]

&$.05$
& $\ms{71.60}{0.44}$ & $\ms{69.32}{0.81}$ & $\ms{68.36}{1.06}$ 
& $\ms{62.07}{0.91}$ & $\ms{62.38}{1.76}$ & $\ms{65.68}{1.28}$ 
& $\ms{12.04}{1.32}$ & $\ms{12.18}{1.69}$ & $\ms{35.95}{2.69}$ 
\\[3pt]

&$0.1$
& $\ms{71.06}{0.34}$ & $\ms{68.70}{0.87}$ & $\ms{67.77}{0.65}$ 
& $\ms{61.53}{1.15}$ & $\ms{61.81}{1.40}$ & $\ms{64.72}{1.08}$ 
& $\ms{12.61}{1.16}$ & $\ms{16.09}{0.93}$ & $\ms{44.60}{4.82}$ 
\\[3pt]

&$0.2$
& $\ms{70.83}{0.38}$ & $\ms{68.04}{0.80}$ & $\ms{66.88}{0.72}$ 
& $\ms{60.93}{0.97}$ & $\ms{59.31}{1.90}$ & $\ms{62.13}{0.84}$ 
& $\ms{13.01}{1.48}$ & $\ms{14.11}{1.51}$ & $\ms{38.54}{4.64}$ 
\\[3pt]

&$0.5$
& $\ms{69.97}{0.45}$ & $\ms{67.11}{1.00}$ & $\ms{65.97}{0.93}$ 
& $\ms{60.08}{1.15}$ & $\ms{58.89}{0.84}$ & $\ms{62.10}{1.02}$ 
& $\ms{12.50}{1.40}$ & $\ms{14.67}{0.78}$ & $\ms{35.44}{3.30}$ 
\\[3pt]

&$1.0$
& $\ms{69.18}{0.43}$ & $\ms{66.42}{0.75}$ & $\ms{65.83}{0.51}$ 
& $\ms{60.01}{1.03}$ & $\ms{57.33}{1.83}$ & $\ms{61.69}{0.98}$ 
& $\ms{12.45}{1.18}$ & $\ms{16.16}{1.81}$ & $\ms{35.11}{3.31}$ 
\\[3pt]

&$2.0$
& $\ms{67.68}{0.60}$ & $\ms{64.82}{0.92}$ & $\ms{64.06}{0.93}$ 
& $\ms{58.14}{1.02}$ & $\ms{56.06}{1.13}$ & $\ms{59.61}{0.50}$ 
& $\ms{12.65}{1.30}$ & $\ms{17.70}{1.79}$ & $\ms{32.96}{3.12}$ 
\\[3pt]

% \midrule
\midrule
\multirow{7}{*}{\begin{turn}{90} CIFAR-10~\texttt{Dir$(0.3)$~} \hfill \end{turn}}
&$0.0$
& $\ms{73.82}{0.37}$ & $\ms{69.28}{0.69}$ & $\ms{71.72}{0.76}$ 
& $\ms{66.12}{1.17}$ & $\ms{43.18}{7.94}$ & $\ms{56.56}{7.78}$ 
& $\ms{8.98}{1.58}$ & $\ms{13.11}{2.26}$ & $\ms{9.87}{0.97}$ 
\\[3pt]

&$.05$
& $\ms{74.10}{0.11}$ & $\ms{72.85}{0.34}$ & $\ms{71.63}{0.54}$ 
& $\ms{66.32}{1.11}$ & $\ms{67.34}{0.91}$ & $\ms{70.27}{0.46}$ 
& $\ms{9.27}{1.39}$ & $\ms{10.94}{3.04}$ & $\ms{53.58}{5.77}$ 
\\[3pt]

&$0.1$
& $\ms{73.53}{0.10}$ & $\ms{72.03}{0.20}$ & $\ms{71.53}{0.41}$ 
& $\ms{65.49}{0.76}$ & $\ms{66.75}{0.52}$ & $\ms{69.88}{0.52}$ 
& $\ms{9.89}{1.34}$ & $\ms{12.37}{2.28}$ & $\ms{52.89}{6.79}$ 
\\[3pt]

&$0.2$
& $\ms{72.89}{0.09}$ & $\ms{71.35}{0.35}$ & $\ms{70.93}{0.25}$ 
& $\ms{64.68}{0.95}$ & $\ms{64.49}{0.96}$ & $\ms{67.97}{0.76}$ 
& $\ms{10.14}{1.14}$ & $\ms{11.60}{1.81}$ & $\ms{42.74}{8.24}$ 
\\[3pt]

&$0.5$
& $\ms{72.60}{0.14}$ & $\ms{70.70}{0.50}$ & $\ms{70.38}{0.35}$ 
& $\ms{64.20}{0.82}$ & $\ms{62.66}{1.67}$ & $\ms{66.97}{0.57}$ 
& $\ms{10.06}{1.20}$ & $\ms{12.65}{1.36}$ & $\ms{43.26}{5.36}$ 
\\[3pt]

&$1.0$
& $\ms{72.11}{0.07}$ & $\ms{70.13}{0.66}$ & $\ms{69.52}{0.49}$ 
& $\ms{64.18}{0.72}$ & $\ms{62.42}{1.44}$ & $\ms{66.49}{0.40}$ 
& $\ms{9.77}{1.44}$ & $\ms{12.80}{1.84}$ & $\ms{39.27}{6.68}$ 
\\[3pt]

&$2.0$
& $\ms{70.71}{0.11}$ & $\ms{69.02}{0.45}$ & $\ms{67.97}{0.58}$ 
& $\ms{63.21}{0.80}$ & $\ms{61.84}{1.11}$ & $\ms{64.61}{0.60}$ 
& $\ms{9.62}{1.36}$ & $\ms{14.43}{3.44}$ & $\ms{39.82}{4.98}$ 
\\[3pt]

\bottomrule
\end{tabular}%
}
% \end{sc}
% \end{small}
\end{center}
\vskip -0.1in
\end{table*}

\subsubsection{Implementation and Evaluation} 
For simplicity, we partition training data roughly evenly among $N$ clients but control their non-IIDness by selecting the Dirichlet distribution parameter in $\{0.1, 0.3\}$; here $N=450$ for EMNIST and $N=1000$ for CIFAR-10. 
% \red{TODO: include one scenario for imbalanced data sizes?} 
In each round, only $S=20$ clients are sampled (uniformly without replacement). 
Honest clients sampled in each round train their local model for $E_c=2$ epochs on their  local data with batch size $B$, where $B=50$ for EMNIST and $B=25$ for CIFAR-10. The server also updates its model for $E_0=2$ epochs in every round using batch size $B_0=180$ (i.e., 10 local steps per round). Additionally, for local training, we use cross-entropy loss with a weight decay of $10^{-4}$ for EMNIST and $10^{-3}$ for CIFAR-10. 
We employ the usual SGD method with a fixed learning rate $\eta=0.1$ in all cases. The number of global rounds is set to be 500 for EMNIST and 1500 for CIFAR-10. 

For our robust aggregation step, we use the following parameters as default: $\alpha=0$ for $\texttt{AF}_{\alpha}$, $(\rho, \theta) = (0.1, 0.5)$ for $\texttt{LF}_{\rho, \theta}$, and $\tau=1$ for $\texttt{Clip}_{\tau}$. For \texttt{GeoMed}, we use the smooth version of Weiszfeld's algorithm in \cite{pillutla2022GeoMed} with a maximum of 4 steps and a relative objective error of $10^{-6}$ as stopping criteria. 

Finally, we use accuracy as the main  metric as test data is balanced. 
Reported numbers are the averages of 3 runs and a rolling window of size 20 for EMNIST and 50 for CIFAR-10.

\subsection{Results}
We first verify that training on the server data alone does not provide significant benefits by using a grid search with a bach size of $\{100, 200\}$, epoch $\{200, 400\}$, learning rate $\{0.002, 0.005, 0.01, 0.1\}$, and optimizer $\{\texttt{SGD}, \texttt{Adam}\}$. The highest test accuracy is roughly $22\%$ for both datasets. In all experiments, we ignore this pretrained model in order to avoid potential overfitting on server data, start with a random model and use \texttt{SGD} exclusively. To study the effect of SF and SL, we vary the SL parameter $\gamma \in \{0, 0.05, 0.1, 0.2, 0.5, 1, 2\}$, the fraction of malicious clients $\beta \in \{0, 0.3, 0.6\}$ for both datasets. The results are given in Table~\ref{table_acc_emnist} above. %and \ref{table_acc_cifar} below. 

% Here is the BIG TABLE
% \input{results_tables}

\begin{figure*}[!tb]
    \centering
    \includegraphics[width=0.333\linewidth]{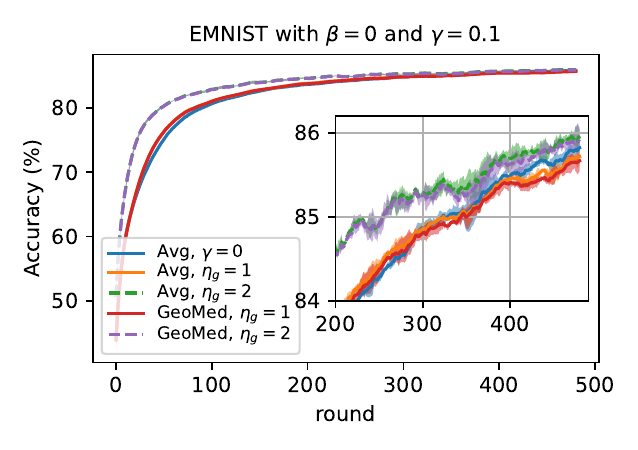}
    \hspace{-4mm}
    \includegraphics[width=0.333\linewidth]{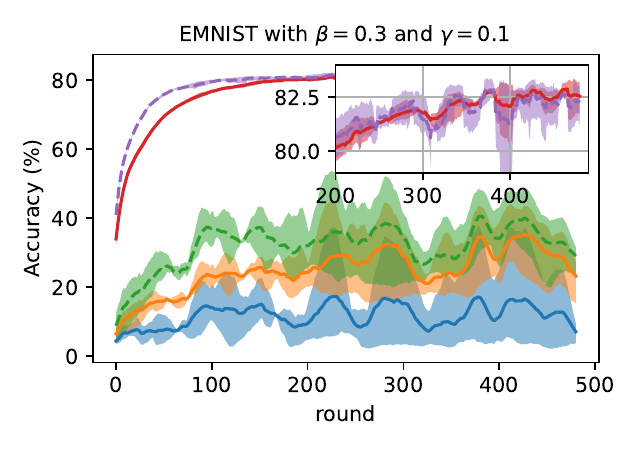}
    \hspace{-4mm}
    \includegraphics[width=0.333\linewidth]{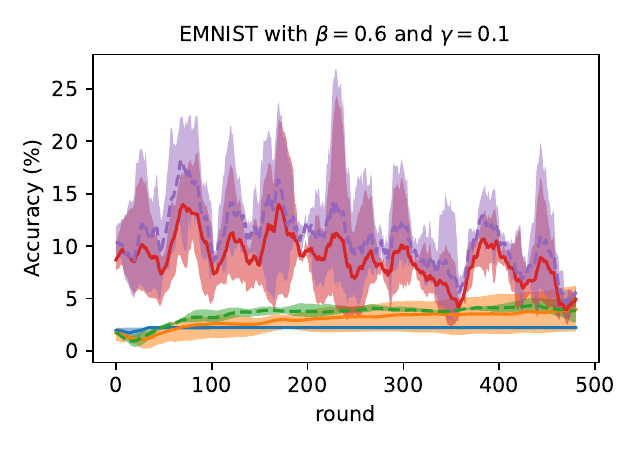}
    \vspace{-3mm}
    \caption{Average accuracy during training on EMNIST-\texttt{Dir}$(0.3)$ using either \texttt{Avg} or \texttt{GeoMed} as the model aggregation step and \textbf{without filtering}. The case without SL is labeled as $\texttt{Avg}, \gamma =0$. We vary the value of $\eta_g \in\{1,2\}$ when using SL with $\gamma=0.1$. Shaded areas represent min-max values over 3 runs.}
    \label{fig_Acc_eta}
    \vspace{-4mm}
\end{figure*}

% \vspace{3pt}
% \noindent \textbf{SL without SF}\\
\subsubsection{SL without server filtering (SF)}
Without attacks, i.e., $\beta=0$, adding SL does not seem to increase the final accuracy in this case. This is because of the following two reasons. First, the server data is rather different from the clients' training data. Second, as mentioned earlier, unlike FSL that adopts a ``pseudo-gradient" step with a step size of $\eta_g>1$ using the aggregated update from clients before the server learning steps \cite{mai2024study}, here we ignore this step (by setting $\eta_g = 1$) throughout our experiments because we believe this step can potentially amplify the effect of malicious updates and reduce the robustness of the algorithm. Fig.~\ref{fig_Acc_eta} confirms this intuition, where SL with $\eta_g =2$ helps accelerate convergence at the beginning and a slight improvement in the final accuracy when $\beta=0$. However, such improvement is not significant when $\beta=0.3$ or $\beta=0.6$, and, in fact, the variation in the final accuracy is actually larger compared to the case with $\eta_g =1$. Thus, in all other experiments, we keep $\eta_g=1$.

% \vspace{6pt}
% \noindent\textbf{SF without SL}\\ 
\subsubsection{SF without SL}
\begin{figure}[t]
    \centering
    \includegraphics[trim = 0 15 0 0, clip, width=0.49\linewidth]{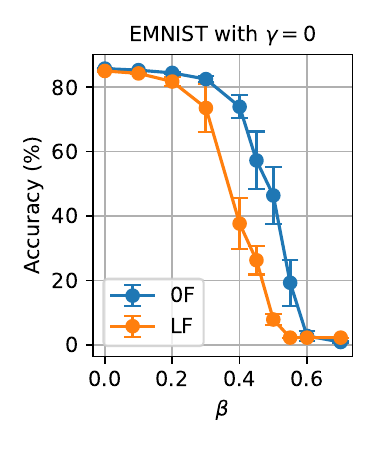}
    \includegraphics[trim = 0 15 0 0, clip, width=0.49\linewidth]{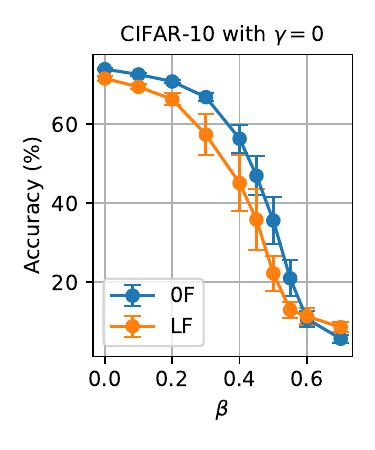}
    \vspace{-5mm}
    \caption{Accuracy vs. malicious fraction $\beta$ when using \texttt{Geometric Median} for model aggregation but \textbf{without SL}.} 
    \label{fig_sensitivity_beta_gam0}
    \vspace{-3mm}
\end{figure}

This corresponds to the rows with $\gamma{=}0$ in Table~\ref{table_acc_emnist}. In this case, except for EMNIST with $\beta{=}0$, adding filters (either \texttt{AF} or \texttt{LF}) can reduce the accuracy considerably compared to \texttt{0F} for both datasets with $\beta=0$ or $\beta=0.3$. To support this observation, we show in Fig.~\ref{fig_sensitivity_beta_gam0} the performance comparison for a finer range of $\beta \in [0,0.7]$. 

This may seem rather counter-intuitive as one often expects the filtering step, when tuned appropriately, to help with reducing the number of corrupted updates for the aggregation step, thereby improving the aggregated model. We attribute this effect to the \texttt{GeoMed}-aggregation step, which is known to have a break point of $0.5$, i.e., it can provide a robust estimator when up to half of the data points are corrupted \cite{lopuhaa1991breakdown}. As a result, when $\beta$ is small, adding the filtering steps (without any finetuning) will likely remove updates from some honest clients (besides some malicious updates, especially those with significant attack strengths), leading to certain degradation in the geometric median of the remaining updates.

\begin{figure}[t]
    \centering
    \includegraphics[trim = 0 12 0 0, clip, width=0.50\linewidth]{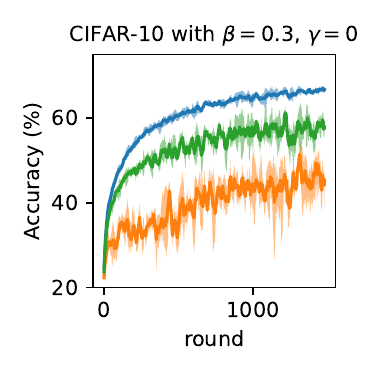}\hspace{-0.1in}
    \includegraphics[trim = 0 12 0 0, clip, width=0.5\linewidth]{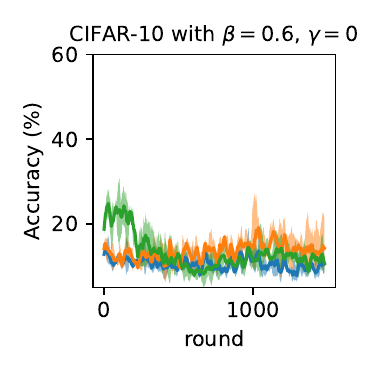}
    \vspace{-2mm}
    \caption{Test accuracy vs. training round on CIFAR-10 Dir$(0.3)$ using \texttt{GeoMed} aggregation, \textbf{without SL}, and with \texttt{0F} (blue), \texttt{AF} (orange), \texttt{LF} (green).}
    \label{fig_acc_gam_cifar10_gam0}
    \vspace{-3mm}
\end{figure}

When $\beta=0.6$, although \texttt{AF} appears to achieve slightly higher accuracy compared to \texttt{0F} and \texttt{LF}, the performance of all schemes without SL degrades significantly. This is because the overwhelming presence of malicious clients sampled in most rounds makes it harder to filter out malicious updates and, as a result, many undesirable updates are used in the model aggregation steps, leading to poor performance. Fig.~\ref{fig_acc_gam_cifar10_gam0} shows the accuracy of the algorithm for CIFAR-10 \texttt{Dir}$(0.3)$. 

\begin{figure}[!t]
    \centering
    \includegraphics[trim = 0 15 0 0, clip, width=0.5\linewidth]{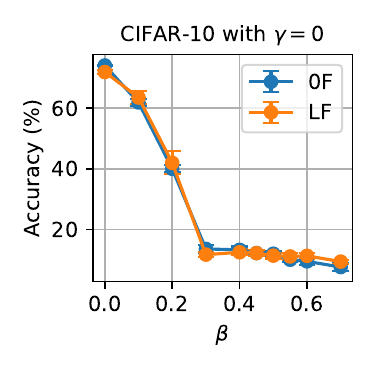}
    \vspace{-2 mm}
    \caption{Accuracy vs. malicious fraction $\beta$ for different filtering and learning parameters at the server and using \texttt{Averaging} aggregation and \textbf{without SL}.} 
    \label{fig_sensitivity_beta_Avg}
    \vspace{-3mm}
\end{figure}

To further illustrate the effect of \texttt{GeoMed} in the case without SL, we show in Fig.~\ref{fig_sensitivity_beta_Avg} the performance of the algorithm when using \texttt{Averaging} for model aggregation instead of \texttt{GeoMed}. In this case, \texttt{LF} without SL is similar to Zeno++\cite{xie2020zeno++}. First, when $\beta=0$, the final accuracy is more or less the same as in Table~\ref{table_acc_emnist} for \texttt{GeoMed}. Also, the filtering step with $\texttt{LF}_{0.1, 0.5}$ reduces the number of updates in half, resulting in lower performance (slower convergence) compared to without filtering \texttt{0F}. Second, as $\beta$ increases, the performance drops significantly compared to the case with \texttt{GeoMed}; e.g., the accuracy is only around $11-12\%$ when $\beta=0.3$ compared to $66\%$ for \texttt{0F} and $56\%$ for \texttt{LF} in Table~\ref{table_acc_emnist}. Here, for small values of  $\beta=0.1, 0.2$ or when $\beta >0.5$, the \texttt{LF} step can provide slight benefits over \texttt{0F}. \vspace{6pt}

% \vspace{6pt}
% \noindent\textbf{Combining SL and SF}\\
\subsubsection{Combining SL and SF}
\begin{figure}
    \centering
    \includegraphics[trim = 0 12 0 0, clip, width=0.50\linewidth]{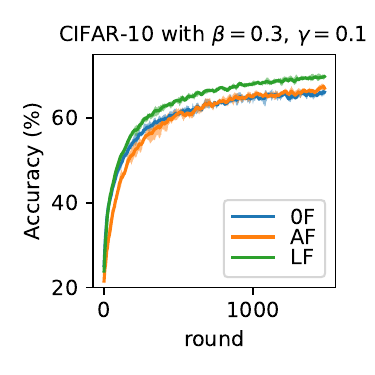}\hspace{-0.1in}
    \includegraphics[trim = 0 12 0 0, clip, width=0.5\linewidth]{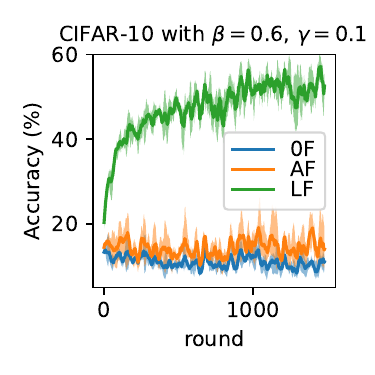}
    \vspace{-2mm}
    \caption{Test accuracy during training on CIFAR-10 Dirichlet 0.3 using \texttt{GeoMed} aggregation with $\gamma=0.1$ for SL.}
    \label{fig_acc_gam_cifar10}
    \vspace{-3mm}
\end{figure}

As we have seen earlier, either SF or SL alone is not enough to counter malicious updates even when using \texttt{GeoMed} for model aggregation. However, when combined together, significant improvement can be achieved for large $\beta$ as shown in Table~\ref{table_acc_emnist}. Not surprisingly, we must select sufficiently large $\gamma$ for SL to take effect. As shown in Table~\ref{table_acc_emnist}, the values of $\gamma$ between $[0.05, 1.0]$ seem to be effective, suggesting that tuning this value need not be difficult in practice. 
Fig.~\ref{fig_acc_gam_cifar10} shows the test accuracy when training on CIFAR-10 Dirichlet 0.3 using SL with $\gamma=0.1$. When $\beta=0.3$, 
%\texttt{AF} is slightly better than \texttt{0F}, 
\texttt{LF} provides significant improvement in terms of convergence rate and final accuracy. In fact, \texttt{LF} works even in the case with $\beta=0.6$ (while \texttt{0F} fails to achieve any learning), suggesting improved robustness of our algorithm. 

It is interesting to note that in the case of $\texttt{AF}_0$, certain values of $\gamma$ and Dirichlet parameter can cause a large drop in the accuracy (e.g., $\gamma = 0.5, 1$ in EMNIST-Dirichlet 0.3) as learning stops when the updated model from the server falls into a local \textit{trap} of server's loss function, where updates from sampled clients are filtered out after a while (explaining 0's in some reported deviations). This does not happen to $\texttt{LF}_{0.1, 0.5}$ as we fix the fraction of filtered updates. Because of such phenomenon and the superior performance of $\texttt{LF}$ over $\texttt{AF}$, we will focus on $\texttt{LF}$ below. 

\begin{figure}
    \centering
    \includegraphics[trim = 0 15 0 0, clip, width=0.48\linewidth]{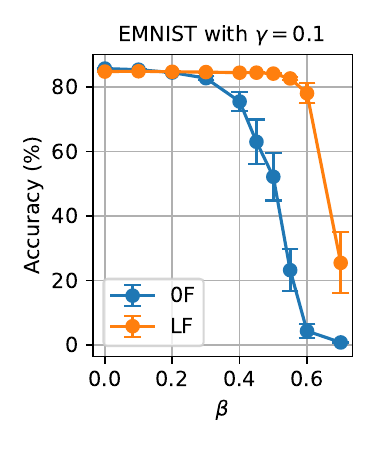}
    \includegraphics[trim = 0 15 0 0, clip, width=0.48\linewidth]{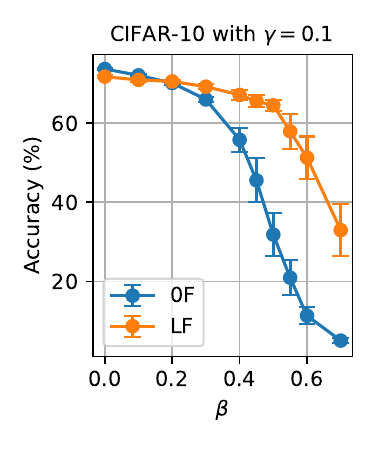}
    \vspace{-2mm}
    \caption{Accuracy vs. malicious fraction $\beta$ for different filtering and learning parameters at the server with  \texttt{Geometric Median} as  model aggregation.}
    \label{fig_sensitivity_beta}
    \vspace{-3mm}
\end{figure}

Next, we vary the fraction of malicious clients for a fixed value of SL to see how the performance degrades. Fig.~\ref{fig_sensitivity_beta} shows the final accuracy of \textsc{RoFSL} for $\beta \in [0,0.7]$, demonstrating a significant improvement in robustness of \texttt{LF} for both datasets.  

% \vspace{6pt}
% \noindent\textbf{Effect of Filter Parameters}\\
\subsubsection{Effect of Filter Parameters} 
\begin{figure}[t]
    \centering
    \includegraphics[trim = 0 15 0 0, clip, width=0.49\linewidth]{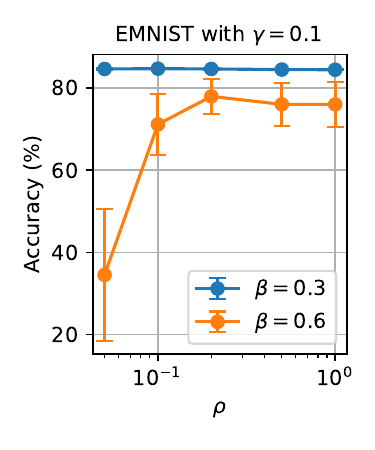}
    \includegraphics[trim = 0 15 0 0, clip, width=0.49\linewidth]{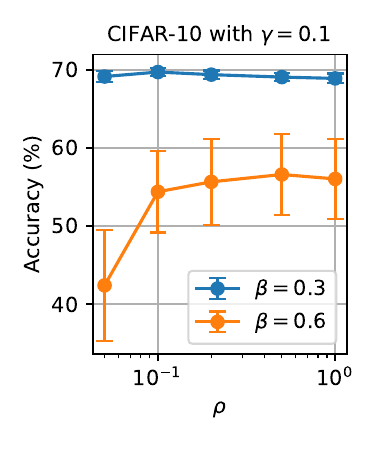}
    \vspace{-5 mm}
    \caption{Sensitivity against the parameter $\rho$ of the loss-based filter \texttt{LF}.}
    \label{fig_sensitivity_rho}
    \vspace{-3mm}
\end{figure}

We focus on the loss-based filter as it shows notable improvements compared to the angle-based filter. Fig.~\ref{fig_sensitivity_rho} shows the sensitivity of the final accuracy against the parameter $\rho$ in $\texttt{LF}_{\rho, \theta}$. For both datasets, the algorithm achieves a consistent performance for the values of $\rho$ between $0.1$ and $1$, suggesting that tuning this parameter should not be a significant challenge in practice. 
% \red{Compare with recent methods? FLAME, FLDetector}

% \vspace{6pt}
% \noindent\textbf{Effect of Non-IIDness}\\
\subsubsection{Non-IIDness}
Table~\ref{table_acc_emnist} shows that in most cases, as expected, the final accuracy reduces as the  client's data become more non-IID. The effects of server learning and filtering is fairly consistent across different datasets and Dirichlet settings. %We will elaborate further below. 
% \vspace{6pt}
% \noindent\textbf{Effect of Client Non-IIDness}

\section{Conclusions and Discussions}\label{sec_conclusion}
We considered auxiliary server learning and filtering as a means to mitigate the performance degradation of FL on non-IID data and Byzantine clients. We showed experimentally that by augmenting the server with a small synthetic dataset and using geometric median for model aggregation, significant improvements in robustness can be achieved. Future work will analyze theoretical guarantee of our algorithm. The  approach here can be viewed as a complementary one that can be extended or combined with other methods. For example, the server may try to estimate the fraction of malicious agents to adaptively adjust filter parameters,  enabling dynamic robustness adaptation. 
Another possibility is to use biased sampling to exclude potentially malicious clients. This can be, for example, based on update consistency like in \cite{zhang2022fldetector}, or the server may track how often a client was sampled but not included in the aggregation scheme and use it as an indicator of potential maliciousness to guide client sampling in later rounds. Such exclusion can help in achieving the objective of optimizing \eqref{eqProblem_SL_Byzantine}, which is over the set of honest clients rather than over \textit{all} clients as in \eqref{eqProblem_SL_normal}.

% Another extension of this work is to consider coordinate attack scehmes where the malicious agents would align their attack vectors to remain stalthy under server's robust filtering. 

% \end{appendices}

\bibliographystyle{ieeetr} %abbrv
\bibliography{ref_full}

\end{document}